\documentclass[
]{ceurart}

\usepackage{listings}
\usepackage{url}
\usepackage{hyperref}
\begin{document}

\copyrightyear{2024}
\copyrightclause{Copyright for this paper by its authors.
  Use permitted under Creative Commons License Attribution 4.0
  International (CC BY 4.0).}

\conference{Forum for Information Retrieval Evaluation, December 12-15, 2024, India}

\title{HateGPT: Unleashing GPT-3.5 Turbo to Combat Hate Speech on X}


\author[1]{Aniket Deroy}[%
orcid=0000-0001-7190-5040,
email=roydanik18@kgpian.iitkgp.ac.in,
]
\cormark[1]
\fnmark[1]
\address[1]{IIT Kharagpur,
  Kharagpur, India}

\author[1]{Subhankar Maity}[%
orcid=0009-0001-1358-9534,
email=subhankar.ai@kgpian.iitkgp.ac.in,
]

\cortext[1]{Corresponding author.}

\begin{abstract}
The widespread use of social media platforms like Twitter and Facebook has enabled people of all ages to share their thoughts and experiences, leading to an immense accumulation of user-generated content. However, alongside the benefits, these platforms also face the challenge of managing hate speech and offensive content, which can undermine rational discourse and threaten democratic values. As a result, there is a growing need for automated methods to detect and mitigate such content, especially given the complexity of conversations that may require contextual analysis across multiple languages, including code-mixed languages. We participated in the English task where we have to classify English tweets into two categories namely Hate and Offensive and Non Hate-Offensive. In this work, we experiment with state-of-the-art large language models like GPT-3.5 Turbo via prompting to classify tweets into Hate and Offensive or Non Hate-Offensive. We modified the temperature as an experimental parameter. In this study, we evaluate the performance of a classification model using Macro-F1 scores across three distinct runs. The Macro-F1 score, which balances precision and recall across all classes, is used as the primary metric for model evaluation. The scores obtained are 0.756 for run 1, 0.751 for run 2, and 0.754 for run 3, indicating a high level of performance with minimal variance among the runs. The results suggest that the model consistently performs well in terms of precision and recall, with run 1 showing the highest performance. These findings highlight the robustness and reliability of the model across different runs.

\end{abstract}

\begin{keywords}
  GPT \sep
    Hate Speech \sep
  Classification \sep
  English \sep
  Prompt Engineering 
\end{keywords}

\maketitle

\section{Introduction}
The advent of social media platforms such as Twitter  (currently known as X) and Facebook has revolutionized the way individuals communicate, enabling people from diverse backgrounds to share their thoughts, experiences, and opinions freely \cite{b1}. This democratization of content creation has led to an exponential increase in user-generated data. While these platforms have facilitated global connectivity and discourse, they have also become hotbeds for hate speech \cite{zhu2020hate,jin2024gpt,aluru2020deep,chakravarthi2023overview} and offensive content. Such content not only disrupts meaningful communication but also poses significant threats to social cohesion and democratic values.

Addressing the proliferation of hate speech \cite{zhang2018detecting,liu2019nuli,saleem2017web} on social media is a complex challenge. The nature of online communication, where context and nuance often play a crucial role, makes it difficult to detect offensive language accurately \cite{b2}. This challenge is further compounded by the multilingual nature of online communities, where users frequently employ code-mixed languages such as Hinglish (a mix of Hindi and English), German-English, and Bangla, among others \cite{b3}. As these languages blend cultural and linguistic elements, the task of identifying hate speech becomes even more intricate.

In response to this growing concern, technology companies and social media platforms have begun to invest in automated methods to detect and manage offensive content \cite{zhu2020hate,anzovino2018automatic,DBLP:conf/fire/NDKBPRKPNC23}. The goal is to strike a balance between preserving open and free dialogue while preventing the spread of harmful speech. In this work, we focus on the classification of English tweets into two categories: Hate and Offensive and Non Hate-Offensive. By leveraging state-of-the-art large language models such as GPT-3.5 Turbo, we experiment with prompting techniques to classify tweets accurately.

Run 1 achieved the highest Macro-F1 score at 0.756, indicating it balanced precision and recall across different classes better than the other runs. Run 2 had a slightly lower score of 0.751, suggesting a small decline in performance, either in precision, recall, or both. Run 3 scored 0.754, which was slightly lower than Run 1 but higher than Run 2, indicating its performance was similar to Run 2's with a minor improvement.

\section{Related Work}

The detection of hate speech and offensive content on social media has garnered significant attention in recent years, driven by the growing need to maintain safe and constructive online environments \cite{b4}. Researchers have explored a variety of approaches to address this issue, ranging from traditional machine learning techniques \cite{smith2018hate, nobata2016abusive} to the application of advanced deep learning models \cite{mozafari2020bert, zhu2020hate}.

Early approaches to hate speech detection \cite{smith2018hate} used simple machine learning algorithms. These models \cite{nobata2016abusive} used manually crafted features, including word n-grams, part-of-speech tags, and sentiment scores, to classify text. For instance, \citet{badjatiya2017deep,chiu2021detecting} employed a logistic regression model with n-grams and part-of-speech features to classify tweets into hate speech, offensive language, and neither. However, the performance of these models was often limited by their reliance on surface-level features, which could not fully capture the complexities of language and context.

\citet{zhang2018detecting} experimented with LSTM networks and Gradient Boosted Decision Trees (GBDT) to classify hate speech on Twitter, demonstrating improvements over traditional machine learning methods. Similarly, \citet{liu2019nuli} utilized a CNN-LSTM architecture to detect offensive language, showing that deep learning models could capture both local and sequential patterns in text.

\citet{mozafari2020bert,saleem2017web} marked the advancement in the field of BERT and transformer. These models, pre-trained on large corpora, allowed for contextual understanding of text, leading to more accurate classification. \citet{mozafari2020hate,zhu2020hate} leveraged BERT for hate speech detection, achieving state-of-the-art performance by fine-tuning the model on labeled datasets. The success of transformer models paved the way for further research into leveraging large language models for offensive language detection.

More recently, the focus has shifted toward leveraging even more sophisticated large language models (LLMs), such as GPT-3 and its successors. These LLMs, with their ability to generate and understand text in a nuanced manner, have shown promise in detecting offensive content. For instance, \citet{chiu2021detecting,mozafari2022cross,thapliyal2020sarcasm} explored the use of GPT-3 for hate speech detection through few-shot learning, highlighting the model’s ability to generalize across different datasets with minimal task-specific training. However, challenges remain in applying these models to code-mixed languages and in ensuring that they can handle the subtleties of context-dependent hate speech.

\citet{yadav2024leveraging,thapliyal2020sarcasm} investigated the detection of hate speech in Hinglish using deep learning models, while  organized a shared task on multilingual hate speech detection \cite{aluru2020deep}, focusing on languages such as English, German, and Hindi. These studies underscore the importance of developing language-agnostic models or approaches that can effectively deal with code-mixing and multilingual content. However, no work has explored the capabilities of GPT-3.5 Turbo for hate speech detection. In this work, we explored the capabilities of GPT-3.5 Turbo to detect hate speech in English social media posts on X.


\section{Dataset}

The English testing dataset consists of 888 tweets collected from a popular social media platform, X. Since we used effectively only the test dataset for our predictions, we only mentioned the statistics corresponding to the test dataset.

\section{Task Definition}
The task~\cite{hasoc_2024_overview_acm,hasoc_2024_overview_ceur} in this study involves the automated classification of social media content, specifically tweets, into two distinct categories: Hate and Offensive (i.e., HOF) and Non Hate-Offensive (i.e., NOT). The objective is to develop a model that can accurately identify whether a given tweet contains hate speech or offensive language (i.e., HOF), or if it does not (i.e., NOT).

\section{Methodology}

Prompting, especially with large language models such as GPT-3.5 Turbo, offers a powerful approach to solving the problem of hate speech and offensive content detection for several reasons:

\begin{itemize}[-]
    
\item \textbf{Contextual Understanding:}
Large language models are pretrained \cite{chen2024large} on vast amounts of text data, enabling them to understand language nuances, context, and semantic relationships between words and phrases. This deep understanding allows them to discern whether a piece of content is offensive or hateful, even when the language is subtle or context-dependent.

\item \textbf{Flexibility and Adaptability:}
Prompting allows for flexibility \cite{dillenbourg2007flexibility} in how the task is framed and tackled. By carefully designing prompts, the model can be directed to focus on specific aspects of the content, such as detecting harmful language or distinguishing between different forms of offense. This adaptability is crucial in handling the diverse and evolving nature of hate speech on social media.


\item \textbf{Multilingual and Code-Mixed Language Handling:}
Prompting large language models is beneficial for dealing with multilingual content \cite{shanmugavadivel2022deep}, including code-mixed languages, which are common on social media. The model’s extensive training on diverse text sources helps it understand and classify content that blends languages or uses non-standard linguistic forms.

\item \textbf{Efficiency in Deployment:}
Prompting does not require the traditional pipeline of data preprocessing \cite{heit2016architecture}, feature extraction, and model training. Instead, the model can be used directly to classify content by providing it with well-crafted prompts. This reduces the time and resources needed to deploy hate speech detection systems.

\item \textbf{Scalability:}
With prompting, the same model can be applied to a wide range of tasks without significant modifications. This scalability \cite{lester2021power} is important for social media platforms that need to monitor vast amounts of content in real time and across different languages.

\item \textbf{Handling Ambiguity and Subjectivity:}
Hate speech and offensive content often involve subjective judgments \cite{curry2024subjective}. Prompting a large language model allows for more nuanced decision-making, as the model can consider context, intent, and the subtleties of language that might be missed by simpler models.

\item \textbf{Rapid Iteration and Improvement:}
Prompting \cite{madaan2024self} enables quick adjustments and refinements based on feedback, making it easier to improve the model’s performance over time. As new forms of offensive language emerge, the prompts can be updated or refined to ensure the model remains effective.

\end{itemize}

\subsection{Prompt Engineering-Based Approach}
We used the GPT-3.5 Turbo\footnote{\url{https://platform.openai.com/docs/models/gpt-3-5-turbo}} model \cite{brown2020language} via prompting to solve the classification task.
We used GPT-3.5 Turbo in zero-Shot mode via prompting.
After the prompt is provided to the LLM, the following steps take place inside the LLM while generating the output. The following outlines the steps that occur internally within the LLM, summarizing the prompting approach using GPT-3.5 Turbo:\\

\textbf{Step 1: Tokenization}

\begin{itemize}
    \item \textbf{Prompt:} \( X = [x_1, x_2, \dots, x_n] \)
    \item The input text (prompt) is first tokenized into smaller units called tokens. These tokens are often subwords or characters, depending on the model's design.
    \item \textbf{Tokenized Input:} \( T = [t_1, t_2, \dots, t_m] \)
\end{itemize}

\textbf{Step 2: Embedding}

\begin{itemize}
    \item Each token is converted into a high-dimensional vector (embedding) using an embedding matrix \( E \).
    \item \textbf{Embedding Matrix:} \( E \in \mathbb{R}^{|V| \times d} \), where \( |V| \) is the size of the vocabulary and \( d \) is the embedding dimension.
    \item \textbf{Embedded Tokens:} \( T_{\text{emb}} = [E(t_1), E(t_2), \dots, E(t_m)] \)
\end{itemize}

\textbf{Step 3: Positional Encoding}

\begin{itemize}
    \item Since the model processes sequences, it adds positional information to the embeddings to capture the order of tokens.
    \item \textbf{Positional Encoding:} \( P(t_i) \)
    \item \textbf{Input to the Model:} \( Z = T_{\text{emb}} + P \)
\end{itemize}

\textbf{Step 4: Attention Mechanism (Transformer Architecture)}

\begin{itemize}
    \item \textbf{Attention Score Calculation:} The model computes attention scores to determine the importance of each token relative to others in the sequence.
    \item \textbf{Attention Formula:}
    \begin{equation}
    \text{Attention}(Q, K, V) = \text{softmax}\left(\frac{QK^T}{\sqrt{d_k}}\right)V
    \end{equation}
    \item where \( Q \) (query), \( K \) (key), and \( V \) (value) are linear transformations of the input \( Z \).
    \item This attention mechanism is applied multiple times through multi-head attention, allowing the model to focus on different parts of the sequence simultaneously.
\end{itemize}

\textbf{Step 5: Feedforward Neural Networks}

\begin{itemize}
    \item The output of the attention mechanism is passed through feedforward neural networks, which apply non-linear transformations.
    \item \textbf{Feedforward Layer:}
    \begin{equation}
    \text{FFN}(x) = \max(0, xW_1 + b_1)W_2 + b_2
    \end{equation}
    \item where \( W_1, W_2 \) are weight matrices and \( b_1, b_2 \) are biases.
\end{itemize}

\textbf{Step 6: Stacking Layers}

\begin{itemize}
    \item Multiple layers of attention and feedforward networks are stacked, each with its own set of parameters. This forms the "deep" in deep learning.
    \item \textbf{Layer Output:}
    \begin{equation}
    H^{(l)} = \text{LayerNorm}(Z^{(l)} + \text{Attention}(Q^{(l)}, K^{(l)}, V^{(l)}))
    \end{equation}
    \begin{equation}
    Z^{(l+1)} = \text{LayerNorm}(H^{(l)} + \text{FFN}(H^{(l)}))
    \end{equation}
\end{itemize}

\textbf{Step 7: Output Generation}

\begin{itemize}
    \item The final output of the stacked layers is a sequence of vectors.
    \item These vectors are projected back into the token space using a softmax layer to predict the next token or word in the sequence.
    \item \textbf{Softmax Function:}
    \begin{equation}
    P(y_i|X) = \frac{\exp(Z_i)}{\sum_{j=1}^{|V|} \exp(Z_j)}
    \end{equation}
    \item where \( Z_i \) is the logit corresponding to token \( i \) in the vocabulary.
    \item The model generates the next token in the sequence based on the probability distribution, and the process repeats until the end of the output sequence is reached.
\end{itemize}

\textbf{Step 8: Decoding}

\begin{itemize}
    \item The predicted tokens are then decoded back into text, forming the final output.
    \item \textbf{Output Text:} \( Y = [y_1, y_2, \dots, y_k] \)
\end{itemize}



\begin{figure}[h!]
  \centering
  \includegraphics[width=0.63\linewidth]{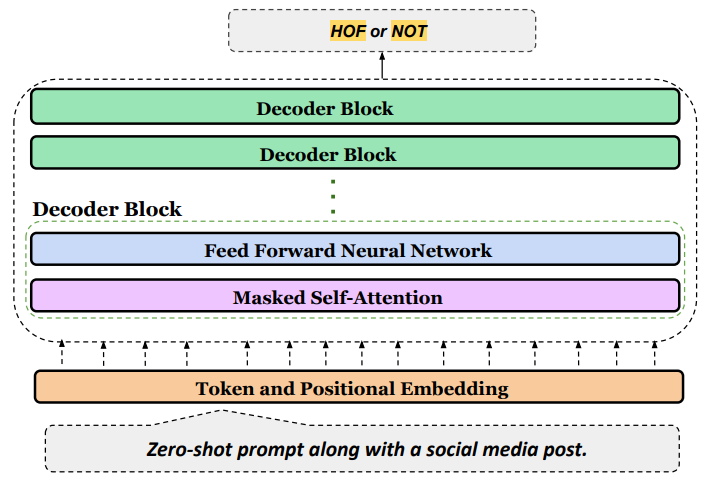}
  \caption{An overview of GPT-3.5 Turbo for hate speech detection using social media posts.} \label{fig2}
\end{figure}

The process begins with tokenization, where the input text is broken down into smaller units called tokens, which could be subwords or characters depending on the model. Next, in the embedding step, each token is converted into a high-dimensional vector using an embedding matrix ,resulting in embedded tokens.
To capture the order of tokens in the sequence, positional encoding is added to the embedded tokens, producing the input for the model. The attention mechanism in the transformer architecture then computes attention scores to determine the importance of each token relative to others. 
Following attention, the output is passed through feedforward neural networks that apply non-linear transformations to enhance the model's learning capacity. The feedforward process involves weight matrices and biases, introducing non-linearity.
These attention and feedforward layers are then stacked to form the deep layers of the model. Each layer processes the input and adds its contribution to the overall understanding of the sequence. The output from the stacked layers is a sequence of vectors.
In output generation, these vectors are projected back into the token space using a softmax layer to predict the next token in the sequence. The softmax function produces a probability distribution over the vocabulary, and the model selects the most likely token. Finally, in decoding, the predicted tokens are converted back into text, forming the final output sequence. This process repeats until the entire output sequence is generated, resulting in the final text produced by the model.

We used the following prompt for English language for the purpose of classification: "\textit{Please Check whether the Tweet-<Tweet> is Hate and Offensive or Non Hate-Offensive. Only state Hate and Offensive or Non Hate-Offensive}".
The figure representing the methodology is shown in Figure~\ref{fig2}.

We run the GPT model at 3 different temperature values- 0.7, 0.8, and 0.9.

All the labels are converted from Hate and Offensive to HOF and from Non Hate-Offensive to NOT and then submitted.

\section{Results}
\begin{table}[h]
    \centering
    \begin{tabular}{cc}
        \toprule
        \textbf{Run Number} & \textbf{Macro-F1 Score} \\
        \midrule
        Run 1 & 0.756 \\
        Run 2 & 0.751 \\
        Run 3 & 0.754 \\
        \bottomrule
    \end{tabular}
    \caption{Macro-F1 Scores for the Three Runs for Hate Speech Detection in English.}
    \label{tab:macro_f1_scores}
\end{table}
Our Team TextTitans ranked 5th \cite{hasoc_2024_overview_acm,hasoc_2024_overview_ceur}, in the English Task for Hate Speech detection. Table~\ref{tab:macro_f1_scores} shows the Macro-F1 scores for the three runs for Hate Speech Detection in English.

Run 1 (0.756): This run has the highest Macro-F1 score among the three runs. This suggests that the model performed slightly better in balancing precision and recall across different classes compared to the other runs.

Run 2 (0.751): This run’s Macro-F1 score is a bit lower than Run 1. The decrease in score might indicate a slight drop in performance, either in precision, recall, or both, across the classes.

Run 3 (0.754): The score for this run is also slightly lower than Run 1 but higher than Run 2. This suggests performance is similar to Run 2 but with a slight improvement.

The scores are relatively close to each other, indicating that the model's performance across these runs is quite consistent.
Differences between the scores are minor, which might be due to variations in the hyperparameter i.e. temperature. These differences, while small, might still be significant depending on the context of the task or the scale of the evaluation.
In summary, all three runs show strong performance with Macro-F1 scores in the range of approximately 0.75 to 0.76. Run 1 shows the best performance among the three, but the differences are minimal, suggesting that the model's performance is relatively stable across these runs.

\section{Conclusion}
In this study, we explored the application of large language models, specifically GPT-3.5 Turbo, for the task of detecting hate speech and offensive content on social media. The increasing volume and complexity of online communication, especially in multilingual and code-mixed languages, present significant challenges for maintaining a safe and constructive digital environment. Our work focused on classifying English tweets into Hate and Offensive and Non Hate-Offensive categories, while also extending our analysis to other languages.

The Macro-F1 scores across the three runs of the classification model demonstrate strong and consistent performance. With scores of 0.756, 0.751, and 0.754, respectively, the results indicate that the model effectively balances precision and recall across different classes. The slight variations observed among the runs are minimal, reflecting the model's stability and reliability in various testing scenarios. These findings affirm the model's capability to perform well in a balanced manner across all classes, reinforcing its utility in practical applications where class performance consistency is critical. Future work may explore further refinements to enhance performance or investigate additional metrics for a more comprehensive evaluation.

While the results are promising, they also highlight areas for further improvement. The complexity of language, the nuances of context, and the evolving nature of online discourse require continuous refinement of models and approaches. Future research should focus on enhancing the model's ability to handle multilingual and code-mixed content more effectively, as well as on developing strategies to address the subjectivity inherent in detecting offensive language.


\bibliography{sample-ceur}


\end{document}